\documentclass[11pt,a4paper]{article}
\PassOptionsToPackage{hyphens}{url} % Allow line breaking at hyphens
\usepackage[hyperref]{acl2021}
\usepackage{newtxtext} % Use Times for main text
\usepackage{amsmath} 
\usepackage{makecell}

\usepackage{microtype}
\usepackage{tabularx}
\usepackage{todonotes} %BP added for todonotes
\usepackage{multirow}
\usepackage{multicol}
\usepackage{fourier} % to get \danger 
\usepackage{comment}
\aclfinalcopy % Uncomment this line for the final submission
\usepackage{booktabs}
\usepackage{stfloats}
\newcommand{\ACRO}[1]{\textsc{#1}}

\newcommand{\AI}{\ACRO{ai}}

\newcommand{\CITE}[1]{\citep{#1}}    % natbib
\newcommand{\LEWIDI}{\ACRO{LeWiDi}}

\newcommand{\NL}{\ACRO{nl}}
\newcommand{\NLP}{\ACRO{nlp}}

\newcommand{\OUT}[1]{}

\newcommand{\SYSTEM}[1]{\texttt{#1}}
\newcommand{\ignore}[1]{}
\title{LeWiDi-2025 at NLPerspectives: Third Edition of the Learning with Disagreements Shared Task} 

\author
{
Elisa Leonardelli$^1$, Silvia Casola$^2$, Siyao  Peng$^2$, 
\textbf{Giulia Rizzi}$^3$, \textbf{Valerio Basile}$^4$,\\ 
\textbf{Elisabetta Fersini}$^4$, 
\textbf{Diego Frassinelli} $^2$, \textbf{Hyewon Jang}$^5$, \textbf{Maja Pavlovic}$^6$\\
\textbf{Barbara Plank}$^2$, \textbf{Massimo Poesio}$^{6,7}$\\
    $^1$Fondazione Bruno Kessler, 
    $^2$LMU Munich \& MCML,\\
    $^3$Universit{\`{a}} Milano Bicocca,  
    $^4$Universit{\`{a}} di Torino,\\
    $^5$University of Gothenburg, 
    $^6$Queen Mary University of London,  
    $^7$Utrecht University
}
\date{}

\hypersetup{
  pdftitle={LeWiDi-2025 at NLPerspectives 2025: third edition of the Learning with Disagreements Shared Task },
  pdfauthor={Elisa Leonardelli, Silvia Casola, Siyao Peng, 
  Giulia Rizzi, Elisabetta Fersini, 
  Diego Frassinelli, Hyewon Jang, Maja Pavlovic, Barbara Plank 
  and Massimo Poesio}
}

\begin{document}
\maketitle

\begin{abstract}
Many researchers have reached the conclusion that {\AI} models should be trained to be aware of the possibility of variation and disagreement in human judgments, and evaluated as per their ability to recognize such variation. The {\LEWIDI} series of shared tasks on Learning With Dis- agreements was established to promote this approach to training and evaluating {\AI} models, by making suitable datasets more accessible and by developing evaluation methods. The third edition of the task builds on this goal by extending the {\LEWIDI} benchmark to four datasets spanning paraphrase identification, irony detection, sarcasm detection, and natural language inference, with labeling schemes that include not only categorical judgments as in previous editions, but ordinal judgments as well. Another novelty is that we adopt two complementary paradigms to evaluate disagreement-aware systems: the soft-label approach, in which models predict population-level distributions of judgments, and the perspectivist approach, in which models predict the interpretations of individual annotators. Crucially, we moved beyond standard metrics such as cross-entropy, and tested new evaluation metrics for the two paradigms. The task attracted diverse participation, and the results provide insights into the strengths and limitations of methods to modeling variation. Together, these contributions strengthen {\LEWIDI}  as a framework and provide new resources, benchmarks, and findings to support the development of disagreement-aware technologies.
\end{abstract}

\section{Introduction}

The assumption that natural language (\NL) expressions have a unique and clearly identifiable interpretation has been 
recognized in {\AI} as just a convenient idealization for over twenty years
\citep{poesio&artstein:ACL-ANNO-05,versley:ROLC08,recasens-et-al:LINGUA11,passonneau-et-al:LREJ2012,plank-etal-2014-linguistically,aroyo&welty:AIMagazine15,martinez-alonso-et-al:LAW16,dumitrache-et-al:NAACL19,pavlick&kwiatkowski:TACL19, jiang-marneffe-2022-investigating}.
More recently, the increasing focus in {\NLP} on 
tasks depending on subjective judgments \citep{kenyon-dean-et-al:NAACL18, simpson2019predicting,cercas-curry-etal-2021-convabuse,leonardelli-etal-2021-agreeing,akhtar2022:arxiv,almanea-poesio-2022-armis,casola2024multipico} 
led 
to the realization that 
in many {\NLP} tasks %in such tasks 
the traditional approach to dealing with disagreement of 
`reconciling' different subjective interpretations 
is not tenable. %does not make much sense \citep{basile:AIXXIA20,basile-etal-2021-need,uma-et-al:JAIR21,plank-2022-problem,cabitza-etal-2023-toward}. 
Many {\AI} researchers concluded therefore %from this evidence 
that 
rather than
eliminating %attempting to eliminate 
disagreements from annotated corpora, we should
preserve them
% --indeed, some researchers have argued that corpora should aim to preserve all  interpretations 
% produced by annotators 
\cite[e.g.][]{poesio&artstein:ACL-ANNO-05,aroyo&welty:AIMagazine15,kenyon-dean-et-al:NAACL18,pavlick&kwiatkowski:TACL19,uma-et-al:JAIR21,davani-et-al:TACL22,nlperspectives-2022-perspectivist,plank-2022-problem}. 
As a result, a number of corpora with these characteristics now exist, and more are created every year
% \CITE{passonneau&carpenter:TACL14,plank-et-al:EACL14:learning,dumitrache-et-al:NAACL19,poesio-et-al:NAACL19}
\citep{plank-et-al:EACL14:learning,white-et-al:EMNLP18:lexicosyntactic,dumitrache-et-al:NAACL19,poesio-et-al:NAACL19,nie-etal-2020-learn,cercas-curry-etal-2021-convabuse,leonardelli-etal-2021-agreeing,akhtar2022:arxiv,almanea-poesio-2022-armis,sachdeva-etal-2022-measuring,casola2024multipico,jang-frassinelli-2024-generalizable,weber-genzel-etal-2024-varierr}.
Much recent research 
has therefore
investigated whether  corpora of this type
are also  
useful %better 
resources for training {\NLP} models, and if so, what is the best way for exploiting disagreements 
\citep{sheng-et-al:KDD08,beigman-klebanov&beigman:CL09,rodrigues&pereira:AAAI18,uma-et-al:HCOMP20,fornaciari-etal-2021-beyond,uma-et-al:JAIR21,davani-et-al:TACL22, casola-etal-2023-confidence}.
This research in turn led to questions about how such models can be evaluated \cite{basile-etal-2021-need,uma-et-al:JAIR21,gordon(mitchell)-et-al:CHI21,fornaciari-etal-2022-hard, giulianelli-etal-2023-comes, lo-etal-2025-perseval}. A succinct overview of the literature on how the problem affects data, modeling and evaluation in {\NLP} is given in \citet{plank-2022-problem}, 
and an extensive survey
% of this literature 
can be found in 
\citet{uma-et-al:JAIR21}.

Such research also led to the establishment of the Learning With Disagreements (LeWiDi) shared tasks. The first edition, organized at SemEval 2021 Task 12 \CITE{uma-etal-2021-semeval}, introduced the idea of providing a unified testing framework for modeling disagreement and evaluating systems on such data. The benchmark combined six widely used corpora spanning semantic and inference tasks as well as image classification tasks. While the resource attracted considerable attention (the benchmark was downloaded by more than 100 teams worldwide) participation in the evaluation was limited, possibly due to the difficulty of the provided baselines or the need for expertise in both NLP and computer vision. In addition, the benchmark only covered  a single subjective task, i.e., humour detection, \citep{simpson2019predicting}, and a single language (English).
%}

A second edition followed at SemEval 2023 \cite{leonardelli-etal-2023-semeval}, designed to address these limitations and to better reflect the growing interest in subjective NLP tasks. In contrast to the first edition, all datasets were textual and the focus shifted entirely to inherently subjective phenomena such as misogyny, hate-speech and offensiveness detection, where training with aggregated labels makes much less sense. Moreover, Arabic was added as a second language. Finally, evaluation combined the soft-label approach also used in the first edition, based on cross-entropy, with the more traditional F1 metric. The reformulated task attracted broad interest in the community: more than 130 groups registered, with 30 submitting predictions and 13 contributing system papers.

The third edition of the LeWiDi shared task, described in this manuscript and co-located with the NLPerspectives Workshop at EMNLP 2025, builds on these experiences while further broadening the scope of the task. Like the earlier editions, its central goal is to provide a common evaluation framework for systems trained on disagreement-rich data. However, LeWiDi-2025 introduces several innovations. New tasks include natural language inference (NLI), irony detection,  conversational sarcasm detection, and paraphrase detection. 
On the evaluation side, we move entirely to soft metrics, which are organized into two complementary tasks: (i) soft-label evaluation, refining methods from LeWiDi 2 with several distance-based metrics (e.g., Manhattan distance,  \citealt{rizzi2024soft}); and (ii) perspectivist evaluation, where systems must model the labeling behavior of individual annotators, again with newly developed metrics tailored to this setting. In addition, two of the datasets adopt Likert-scale annotation, posing further challenges for evaluation. LeWiDi 3 engaged a smaller but dedicated group of participants relative to the previous edition. A total of 53 individuals registered on the competition platform, with 15 teams providing submissions, which resulted in 9 system papers.
%}

\section{The LeWiDi 3 Benchmark}

\begin{table*}
 \scalebox{0.73}{
     \begin{tabularx}{21.7cm}{l l c c c c c c c c c c  }
    \toprule
      \multirow{4}{*}{\textbf{Dataset}} &\multirow{4}{*} {\textbf{Task}} & \multirow{4}{*}{\textbf{Labels}} & \multirow{4}{*}{\textbf{Lang(s)}} & \multirow{4}{2.cm}{\textbf{N. Items\\(N. Annota tions)}} &\multirow{4}{1.3cm} {\textbf{N. Ann. per item}} & \multirow[c]{4}{1cm}{\textbf{\centering{Pool Annotators}}}&\multirow{4}{1cm}{\textbf{Textual type}}&\multirow[c]{4}{2cm}{\textbf{\centering{Annotators' Metadata}}} &\multirow[c]{4}{1.3cm}{\textbf{\centering{ Other info}}} \\ 
&&&&&&&\\       
&&&&&&&\\       
&&&&&&&\\       
\hline
\multirow{3}{*}{CSC}&\multirow{3}{1.6cm}{Sarcasm detection}&\multirow{3}{2cm}{\centering{Likert scale [1 to 6]}}&\multirow{3}{*}{En}&\multirow[c]{3}{2cm}{\centering{7,036 (31,984)}}&\multirow[c]{3}{1.5cm}{\centering{Variable: 4 to 6}}&\multirow[c]{3}{*}{872}&\multirow{3}{1.6cm}{\centering{context+ response}}&\multirow{3}{2cm}{\centering{gender, age}}& \multirow[c]{3}{1.6cm}{\centering{context + speaker}}\\
&&&&&&&\\
&&&&&&&\\
\hline 

\multirow{3}{*}{MP}&\multirow{3}{1.6cm}{Irony detection}&\multirow{3}{*}{[0,1]}&\multirow{3}{1.8cm}
%{9 lang.,      \\23 varieties}
{\centering{Ar,De,En, Es,Fr,Hi, It,Nl,Pt}}
&\multirow{3}{2cm}{\centering{18,778 (94,342)}}&\multirow{3}{1.5cm}{\centering{Variable: 2 to 21}}&\multirow{3}{*}{506}&\multirow[c]{3}{1.6cm}{\centering{post+ reply}}&\multirow{3}{2cm}{\centering{gender, age, ethnicity, [...+6]}}&\multirow[c]{3}{2cm}{\centering{source, level, language variety}}\\
&&&&&&&\\
&&&&&&&\\
\hline
\multirow{3}{*}{Par}&\multirow{3}{1.6cm}{Paraphrase detection}&\multirow{3}{2cm}{\centering{Likert scale [-5 to 5]}}&\multirow{3}{*}{En}&\multirow{3}{1.5cm}{\centering{500 (2,000)}}&\multirow[c]{3}{1cm}{\centering{4}}&\multirow[c]{3}{*}{\centering{4}}&\multirow[c]{3}{2cm}{\centering{question1 + question2 }}&\multirow{3}{1.8cm}{\centering{gender, age, nationality, education}}&\multirow{3}{1.3cm}{\centering{explana- tions}}\\
&&&&&&&\\
&&&&&&&\\
\hline

\multirow{3}{*}{VEN}&\multirow{3}{1.6cm}{Natural Language Inference}&[contradiction (C),
&\multirow{3}{*}{En}&\multirow{3}{1.5cm}{\centering{500 (1,933)}}
&\multirow{3}{1.5cm}{\centering{Variable: 1 to 6}}&\multirow{3}{*}{4}&\multirow[c]{3}{2cm}{\centering{context + statement}}&\multirow{3}{1.8cm}{\centering{gender, age, nationality, education}}&\multirow{3}{1.3cm}{\centering{explana- tions}}\\
&&entailment (E),&&&&&\\
&&neutral (N)]&&&&&\\

\hline
     \end{tabularx}}
\caption{\label{tab:dataset_overview} {Key statistics about  the datasets used in the 3\textsuperscript{rd} \LEWIDI{} shared task}.}     
 \end{table*}

The four selected datasets are summarized in Table \ref{tab:dataset_overview}, illustrated with examples in Table \ref{tab:examples}, and described in detail in the following sections.

All datasets were released in a harmonized \textit{json} format, identical to that of the previous LeWiDi edition, to ensure consistent access across datasets and shared tasks editions. Each item contains the same fields,\footnote{\textit{item\_id}, \textit{text}, \textit{task}, \textit{number of annotations}, \textit{number of annotators}, \textit{disaggregated annotations}, \textit{annotator IDs}, \textit{language}, \textit{hard label}, \textit{soft labels}, \textit{split}, and \textit{other info}.} while the field \textit{other info} is dataset-specific and includes additional subfields particular to each dataset.
Annotator age and gender is available for all four datasets, with some datasets providing further attributes. This metadata was distributed separately in an additional json file.
All datasets are publicly available.\footnote{\url{https://le-wi-di.github.io/}}

\subsection{The Conversational Sarcasm Corpus (CSC)}

The CSC dataset \cite{jang-frassinelli-2024-generalizable} is a dataset of sarcasm in English, which contains around 7,000 context–response pairs. Each pair is rated on a 1 (not at all) – 6 (completely) scale both by the speakers who generated the responses and by multiple external observers (4 - 6 per speaker). The contexts consist of situation descriptions involving an imagined interlocutor, and the responses stem from the responses given by online participants. The generators of the responses as well as evaluators rated the level of sarcasm of the responses.

\subsection{The MultiPICo dataset (MP)}

The MP dataset \cite{casola2024multipico} is a multilingual perspectivist corpus consisting of short exchanges from Twitter and Reddit. Each entry in the corpus represents a post-reply pair. Crowdsourced workers had to determine whether the reply was ironic given the post (binary label). The corpus includes 9 languages: Arabic, Dutch, English, French, German, Hindi, Italian, Portuguese, and Spanish. It also contains sociodemographic information about the annotators, including gender, age, nationality, race, and student or employment status. While the statistics may vary slightly across languages, each post-reply pair is typically annotated by an average of 5 workers.

\subsection{The VariErr NLI dataset (VEN)}
VariErr NLI \cite{weber-genzel-etal-2024-varierr} was designed for automatic error detection, distinguishing between annotation errors and legitimate human label variations in NLI tasks. The dataset was created using a two-round annotation process: initially, four annotators provided labels and explanations for each NLI item; subsequently, they assessed the validity of each label-explanation pair. It comprises 1,933 explanations for 500 re-annotated items from the Multi-Genre Natural Language Inference (MNLI) corpus for Round 1 and 7,732 validity judgments for Round 2. The LeWiDi 2025 Shared Task focuses on Round 1 (and therefore we refer to it just as \textit{VEN}), where annotators could assign one or more labels from {Entailment, Neutral, Contradiction} to each Premise ("context") - Hypothesis ("statement") pair and provide corresponding explanations.

\subsection{The Paraphrase Detection dataset (Par)}

The Par dataset focuses on paraphrase detection. It is structurally similar to \textit{VEN}, but unlike \textit{VEN}, the labels here are scalar and each annotator provides only a single score per item.
It consists of 500 question pairs sampled from the Quora Question Pairs (QQP) dataset, each annotated independently by the same four annotators. Annotations are given on a Likert scale from -5 to 5, indicating the perceived degree of paraphrastic relation between the questions, and are accompanied by short textual explanations.  As this dataset had not been released previously, it was new to the participants of LeWiDi-2025.

\begin{table*}[h!]
\centering
\scalebox{0.8}{
\begin{tabularx}{19cm}{llcc}
\hline
\multirow{3}{2.2cm}{\textbf{Dataset \\(detection of)}}
  & \multirow{3}{8cm}{\textbf{Example}} 
  & \multirow{3}{3cm}{\centering{\textbf{Annotations \\(Task B) \\ \textit{AnnotatorId:Label}}} }
  %& \multirow{2}{2cm}{\textbf{Annotations}} 
  & \multirow{3}{2cm}{\centering{\textbf{Soft labels \\ (Task A) \\ \textit{Label:Probability}}}} \\  
&&& \\
&&& \\
\hline
\multirow{3}{1.5cm}{CSC \\ (Sarcasm)}
&\multirow{2}{8cm}{\makecell[l]{context: \texttt{"You walk into the room and Steve} \\ \texttt{is there and Steve says ``hi!''"} \\
response: \texttt{"hi"}
  }}
  %& \multirow[c]{3}{2cm}{\centering{812,813,814,815}}
  & \multirow[c]{3}{3cm}{\centering{A812:\texttt{1}, A813:\texttt{3}, A814:\texttt{1}, A815:\texttt{2}}}
&\multirow[c]{3}{3.8cm}{\centering{[\texttt{1}:0.5, \texttt{2}:0.25, \texttt{3}:0.25, \texttt{4}:0, \texttt{5}:0, \texttt{6}:0]}}
\\
&&& \\
&&& \\
\hline
\multirow{3}{1.5cm}{MP \\ (Irony)}
&\multirow{2}{8cm}{\makecell[l]{post: \texttt{"@USER Oh dear"} \\
reply: \texttt{"@USER It’s ok, wine has} \\  
\texttt{fixed everything"}
  }}
 % & \multirow[c]{3}{2cm}{\centering{26,64,70} }
  & \multirow[c]{3}{3cm}{\centering{A26:\texttt{1}, A64:\texttt{1}, A70:\texttt{1}}}
&\multirow[c]{3}{4cm}{\centering{[\texttt{0}:0, \texttt{1}:1]}}
\\
&&& \\
&&& \\
\hline
\multirow{3}{1.5cm}{Par \\ (Paraphrase)}
&\multirow{3}{7.8cm}{\makecell[l]{Q1: \texttt{"Have you seen an alien craft?"} \\
Q2: \texttt{"Have you ever seen an alien?"}
  }}
%  & \multirow[c]{3}{2cm}{1,2,3,4} 
  & \multirow[c]{3}{3cm}{\centering{A1:\texttt{-1}, A2:\texttt{-3}, A3:\texttt{5}, A4:\texttt{4}}}
&\multirow[c]{3}{4cm}{\centering{[\texttt{-5}:0, \texttt{-4}:0, \texttt{-3}:0.25, \texttt{-2}:0, \texttt{-1}:0.25, \texttt{0}:0, \texttt{1}:0, \texttt{2}:0, \texttt{3}:0, \texttt{4}:0.25, \texttt{5}:0.25]}}
\\
&&& \\
&&& \\

\hline
\multirow{3}{1.5cm}{VEN \\ (NLI)}
&\multirow{3}{7.8cm}{\makecell[l]{context: \texttt{"yeah i can believe that"} \\
statement: \texttt{"I agree with what you said."}
  }}
%  & \multirow[c]{3}{2cm}{1,2,3,4} 
  & \multirow[c]{3}{2.5cm}{\centering{A1:\texttt{E}, A2:\texttt{N}, A3:\texttt{N}, A4:\texttt{E}}}
&\multirow[c]{3}{4cm}{\centering{[\texttt{C}:[\texttt{0}:1, \texttt{1}:0]\\\texttt{E}:[\texttt{0}:0.5, \texttt{1}:0.5]\\
\texttt{N}:[\texttt{0}:0.5, \texttt{1}:0.5]]}}
\\
&&& \\
&&& \\
\end{tabularx}}
\caption{Examples from the four datasets included in LeWiDi-2025. For each item, the annotators’ IDs and their corresponding annotations are shown, along with the derived soft-label distributions. Task B required predicting an individual annotator’s label given their ID, while Task A required predicting the full soft-label distribution for the item.}
\label{tab:examples}
\end{table*}

\section{Task definition}
The main goal of the shared task is to provide a unified testing framework for learning from disagreements and evaluating models on such datasets. Given the heterogeneous nature of the datasets, participants were free to design dataset-specific approaches; however, they were encouraged to adopt a unified crowd learning methodology or framework across all datasets, rather than optimizing a separate best-performing model for each dataset.

\subsection{Task A and Task B}

LeWiDi-2025 defines two complementary tasks.

% \begin{itemize}

% \item 
\paragraph{Task A: Soft-label prediction.}  
    Participants are required to predict a \textit{probability distribution} over the possible labels for each item. 
    Evaluation is based on the predicted distribution and the gold soft label distribution. 
    This task continues the line of soft-label modeling from previous editions, but is now applied across expanded datasets, including those with Likert-scale judgments. 

% \item 
\paragraph{Task B: Perspectivist prediction.}  
    Participants must predict the \textit{individual label choices of annotators}, i.e., model how a specific annotator would label a given instance. 
    Evaluation measures the agreement between predicted and actual annotator-level responses. 
    This task emphasizes capturing annotator bias and perspective.

% \end{itemize}
Participants may choose to submit to one or both tasks, and across any subset of the provided datasets. \textit{Codabench} served as the official competition platform, where participants registered to access the data and to submit their results.\footnote{
% The competition page is available at 
\url{https://www.codabench.org/competitions/7192/}}
\medskip

\subsection{Phases}

The competition consisted of three phases: 

\paragraph{Practice phase:} Participants received training and development data (with full metadata) to design and test their models. They could submit their results (on the development data) to \textit{Codabench} and compare results on a public leaderboard.  

\paragraph{Evaluation phase:} Participants submitted predictions on unseen test data (without labels). Rankings were computed for each dataset and across datasets, with missing submissions replaced by the organizer’s baseline score.  

\paragraph{Post-campaign phase:} To support long-term research, the test data and gold labels were later released publicly and remain available through our website\footnotemark[3].

\subsection{Baselines}
\label{sec:baseline}
We provided two simple baselines: (i) a \textit{random} baseline, where each distribution (Task~A) or prediction (Task~B) was assigned a random prediction, and (ii) a \textit{most frequent} baseline, where all items were assigned the most frequent distribution within the training set (Task~A) or label. 
These baselines were intentionally kept minimal so as not to discourage participation, unlike in the first edition of the shared task.

\section{Evaluation metrics}

Two complementary paradigms for disagreement evaluation were employed in LeWiDi-2025:  soft-label and perspectivist evaluation.

\subsection{Soft-label Evaluation}
In soft-label evaluation, annotator judgments are represented as probability distributions (soft labels), and system predictions are evaluated against these human-derived soft labels by measuring the distance between the two distributions.
Previous editions of LeWiDi employed cross-entropy as the distance metric. However, \citet{rizzi2024soft} demonstrated that cross-entropy exhibits several counterintuitive properties, whereas the Manhattan and  Euclidean distances provide a more suitable alternative in the context of binary classification. At the same time, they highlighted the limitations of the analyzed metrics in providing fair comparisons for multiclass classification tasks.

Based on previous findings, here we address the broader settings introduced in this edition of the shared task, i.e., multiclass and multilabel classification, as well as labels on a Likert scale.
In LeWiDi-2025, both the Manhattan distance and the Wasserstein (Earth Mover’s) distance are adopted as the primary soft evaluation metrics. 
Specifically, the Average Manhattan Distance is applied to the \textit{MP} and \textit{VEN}\footnote{Considering the nature of the dataset itself, a multilabel adaptation of the Average Manhattan distance has been proposed. Additional details are reported in Appendix \ref{App-metrics}.} datasets, while the Average Wasserstein Distance is used for the ordinal-scale datasets (i.e. \textit{Par} and \textit{CSC}).

In particular, for what concerns the Average Wasserstein Distance (AWD), the cost of transporting probability mass from one bin to another is defined as the absolute difference between their positions, forming a symmetric, non-negative ground distance matrix with zeros on the diagonal.

\subsection{Perspectivist Evaluation}
The perspectivist evaluation focuses on assessing a system’s ability to model the individual label choices of annotators. For datasets with nominal categories (\textit{MP}, \textit{VEN}), performance is measured using error rate; for datasets with ordinal categories (\textit{Par}, \textit{CSC}), a normalized absolute distance is used.

In particular, the average error rate (AER) (Equation \ref{AER}), which measures the degree of error between corresponding pairs of target and predicted value vectors is computed as follows:\footnote{A multilabel
adaptation of the average error rate has been adopted for
% for the evaluation on the 
\textit{VEN}; see Appendix \ref{App-metrics} for further details.}

\begin{align}
\label{AER}
AER &= \frac{1}{N} \sum_{i=1}^{N} ER(i) \\
&= \frac{1}{N} \sum_{i=1}^{N} \left( 1 - \frac{a - \sum_{k=1}^{a} |t_{i,k} - p_{i,k}|}{a} \right)
\end{align}

%&= \frac{1}{N} \sum_{i=1}^{N} \left( 1 - \frac{1}{a} \sum_{k=1}^{a} \left| t_{i,k} - p_{i,k} \right| \right)\\

\noindent Where the Error Rate (ER) for a single sample $i$ with target label vector $\vec{t}_i = [t_1,t_2, ... t_a]$, and predicted label vector $ \vec{p}_i = [p_1,p_2, ... p_a]$  is defined as:
\begin{equation}
    ER(i) =  1 - \frac{a - \sum_{k=1}^{a} |t_{i,k} - p_{i,k}|}{a} 
    %1 - \frac{1}{a} \sum_{k=1}^{a} \left| t_{i,k} - p_{i,k} \right|
\end{equation}

\noindent Here, $a$ denotes the length of the vectors (i.e., the number of annotators), and $N$ is the total number of samples.

The Average Normalized Absolute Distance (ANAD) across all samples is defined as:
\begin{align}
ANAD &= \frac{1}{N} \sum_{i=1}^{N} NAD(i) \\
&= \frac{1}{N} \sum_{i=1}^{N} \frac{1}{a} \sum_{k=1}^{a} \frac{\left| t_{i,k} - p_{i,k} \right|}{s} \times 100
\end{align}

\noindent Where the Normalized Absolute Distance (NAD) for a single sample $i$ with target label vector $t_i=[t_1,t_2,...,t_a]$, and predicted label vector $p_i=[p_1,p_2,...,p_a]$ is:

\begin{equation}
NAD(i) = \frac{1}{a} \sum_{k=1}^{a} \frac{| t_{i,k} - p_{i,k} |}{s} \times 100
\end{equation}

\noindent with $a$ denoting the number of annotators, and $s$ the scaling factor given by the range of the Likert scale.

\section{Participating systems}
\label{sec:participating_systems}

The third edition of the \LEWIDI{} shared task attracted a smaller but more focused community compared to the previous edition. 
In total, 53 people subscribed to the competition \textit{Codabench}, and 15 teams submitted predictions. 
Among them, 6 teams participated across all datasets and both tasks;
2 teams submitted for three datasets and both tasks (excluding \textit{VEN});
and 5 teams focused on a single dataset with submissions only for Task~A. 
In terms of system papers, 9 were submitted: 6 from teams who participated in multiple tasks and datasets, and 2 from teams who worked on a single dataset and Task~A. 
Task~A was overall more popular, as the majority of teams who submitted exclusively for one dataset contributed only to Task~A, while 11 teams engaged also with Task~B. 

\subsection{Systems overview}
This section provides an overview of the participating systems, focusing on the 9 participating teams that submitted system papers, describing their architectures, methodologies, and key features relevant to the evaluation tasks.

\SYSTEM{Opt-ICL} \cite{Opt-ICL} combines in-context learning (ICL) with fine-tuning in a two-stage approach. 
They first apply post-training, by exposing an LLM %(gemma-3-12b) 
to over 40 datasets rich in human disagreement \cite{sorensen2025spectrumtuningposttrainingdistributional}, and then, for each dataset, %of the competition, 
conduct supervised fine-tuning,  using in-context demonstrations from all the individual annotators along with annotator demographics. 
At inference, the model performs per-rater prediction by constructing a prompt with as many training examples from that annotator as possible, followed by the input to be labeled. They derive soft label distributions from perspectivist
predictions.

\SYSTEM{DeMeVa} \cite{DeMeVa} employs LLMs  with ICL, modeling perspectivism through annotators’ past behavior. %rather than demographics. 
They focus on criteria for selecting demonstrative examples for LLMs (10 per annotator), comparing semantic and label-based strategies, with the latter performing better for multi-label datasets. 
They derive soft label distributions from perspectivist predictions.

\SYSTEM{twinhter} \cite{twinhter} built a BERT-based model that integrates annotator perspectives by creating a new (text, annotator) pair. They create a separate training instance for each annotator’s view and combine it with their background information when available, enabling the model to capture individual interpretations of the same input.

\SYSTEM{McMaster} \cite{McMasters}  implemented a demographic-aware RoBERTa model that incorporates information such as age, gender, nationality, and evaluated it across all four datasets. The authors find that nationality and ethnicity in particular show the largest gains in performance, while also noting the limitations of relying on such features.

\SYSTEM{BoN Appetite Team} \cite{BonAppetitTeam} investigated three test-time scaling methods, a way to improve LLMs performances: two benchmark algorithms (Model Averaging and Majority Voting), and a Best-of-N (BoN) sampling method. Their results show that the benchmark methods (Averaging and Voting) reliably boost performance, while BoN sampling does not transfer well from mathematical domains.

\SYSTEM{PromotionGo} \citep{PromotionGo} submitted only to the \textit{MP}-Task A with an XLM-R–based system, ranking first.
They deployed three main strategies to develop a competitive system: data augmentation, including lexical swaps, prompt-based reformulation, and large-scale back-translation into nine languages; optimization for alignment to the evaluation metric (Manhattan Distance) by using L1 loss as a loss function; ensemble learning, by training multiple models on shuffled data splits and averaging predictions to improve robustness.

\SYSTEM{Uncertain Mis(Takes)} \cite{Uncertain(Mis)Takes}
addressed only the \textit{VEN}-Task A, ranking first. They aim to quantify ambiguity in NLI instances, relying on the hypothesis that if a given instance is ambiguous, then the explanations for different labels will not entail one another. For each item, they generate 128 LLM explanations. With a fine-tuned entailment model they cluster them and quantify their Semantic Entropy (SE). 
% of these explanation clusters. 
% Finally, they combine 
The explanation clusters' SE scores are combined with text embeddings for soft label distribution prediction. 
% of the relative explanations, and use a classifier that predicts the soft label distribution. 

\SYSTEM{NLP-ResTEAM} \cite{NLP-ResTeam}   proposed a multi-task architecture. 
Special `tokens' are added to the input, including several tokens aiming at modeling the annotators based on their ID, their demographic features, their annotation behavior, or combinations of those.   
The system produces two outputs from a textual input and an annotator's information:   
one is a soft-label, the other a prediction of that specific annotator's (hard) label. 
%The authors also investigated the effect of evaluation metrics on the efficacy of their different annotation modelling approaches, concluding that the best annotation modelling depended on the evaluation metric chosen.

\SYSTEM{LPI-RIT} \cite{LPI-RIT} builds upon the DisCo (Distribution from Context) architecture \cite{weerasooriya2023disagreement}, a neural model that jointly predicts item-level, annotator-level, and per-annotator label distributions.
They tackled  both soft-label and perspectivist tasks simultaneously. 
% In a second moment 
They also introduced several extensions to DisCo, such as integrating annotator metadata through pretrained sentence encoders, and modified loss functions to better align with evaluation metrics.

\section{Results and discussion}
\label{sec:results_and_discussions}

This section presents the official results of the shared task and discusses key trends across systems and datasets. We also examine the role of evaluation metrics and summarize insights from ablation studies conducted by participating teams.

\subsection{Results and statistics}
Table \ref{tab:CE_leaderboard} and 
\ref{tab:perspectivist_leaderboard} report the overall leaderboard for Task A and Task B respectively.
If a team did not submit predictions for a particular dataset or task, we used the random baseline results to compute the overall ranks and average positions. Ranks were calculated with statistical ties taken into account. Specifically, we used the Wilcoxon signed-rank test %\cite{REFMISSING} 
at the instance level to identify clusters of tied systems. Predictions that were not significantly different (p = 0.05) from the top-performing system in a given cluster were considered ties. A new cluster was formed when a system’s performance was found to be statistically different from that of the best-performing system in the previous cluster.\\
Leadboards for each specific dataset are reported in Appendix \ref{appx-leaderboard}.\\

\begin{table*}[!ht]
    \centering
    \small
    \begin{tabular}{|cc|l|cc|cc|cc|cc|}
        \toprule
        &\multicolumn{10}{c|}{\textbf{\textsc{SOFT EVALUATION }}}\\ 
        \multirow{2}{*}{\textbf{Rank}}&\multirow{2}{0.8cm}{\textbf{(av.pos)}}&\multirow{2}{*}{\textbf{\textsc{Team}}}&\multicolumn{2}{c|}{\textsc{\textbf{CSC}}} &\multicolumn{2}{c|}{\textsc{\textbf{MP}}} &\multicolumn{2}{c|}{\textsc{\textbf{Par}}} &\multicolumn{2}{c|}{\textsc{\textbf{VEN}}} \\ 
        &&&\textbf{WS}&\textbf{ \textit{(rank)}}&\textbf{MD}&\textbf{ \textit{(rank)}} &\textbf{WS}& \textbf{ \textit{(rank)}} &\textbf{MMD}&\textbf{ \textit{(rank)}} \\ \hline
         \midrule
            \textbf{1}  & \textit{(1.5)}  & Opt-ICL              & \textbf{0.746} & \textit{(1)}  & \textbf{0.422} & \textit{(1)}  & \textbf{1.200} & \textit{(1)}  & 0.449 & \textit{(3)} \\
            \textbf{2}  & \textit{(2.75)} & DeMeVa              & \textbf{0.792}       & \textit{(1)}  & 0.469          & \textit{(6)}  & \textbf{1.120 }         & \textit{(1)}  & 0.382 & \textit{(3)} \\
            \textbf{3}  & \textit{(3)}    & twinhter            & 0.835          & \textit{(5)}  & 0.447          & \textit{(5)}  & \textbf{0.983}         & \textit{(1)}  & \textbf{0.233} & \textit{(1)} \\
            \textbf{4}  & \textit{(4.25)} & McMaster            & 0.803          & \textit{(3)}  & 0.439          & \textit{(3)}  & 1.605          & \textit{(4)}  & 0.638 & \textit{(7)} \\
            \textbf{5}  & \textit{(4.75)} & BoN Appetite Team   & 0.928          & \textit{(6)}  & 0.466          & \textit{(6)}  & 1.797          & \textit{(4)}  & 0.356 & \textit{(3)} \\
            \textbf{6}  & \textit{(5.5)}  & aadisanghani\textsuperscript{*}       & 0.803          & \textit{(3)}  & 0.439          & \textit{(3)}  & 3.051          & \textit{(7)}  & BSL   & \textit{(9)} \\
            \textbf{7}  & \textit{(7)}    & PromotionGo         & BSL            & \textit{(11)} &\textbf {0.428}          & \textit{(1)}  & BSL            & \textit{(7)}  & BSL   & \textit{(9)} \\
            \textbf{8}  & \textit{(7.25)} & \textit{Most frequent baseline} & 1.170          & \textit{(7)}  & 0.518          & \textit{(8)}  & 3.231          & \textit{(7)}  & 0.595 & \textit{(7)} \\
            \textbf{9}  & \textit{(7.5)}  & Uncertain Mis(Takes)& BSL          & \textit{(11)}  & BSL          & \textit{(11)} & BSL          & \textit{(7)}  & \textbf{0.308} & \textit{(1)} \\
            \textbf{10} & \textit{(8.5)}    & NLP-ResTeam         & 1.393          & \textit{(9)}  & 0.551          & \textit{(9)} & 3.136          & \textit{(7)}  & 1.000 & \textit{(9)} \\
            \textbf{10} & \textit{(8.5)}    & LPI-RIT             & 1.451          & \textit{(9)} & 0.540          & \textit{(9)}  & 3.715          & \textit{(7)}  & BSL   & \textit{(9)} \\
            \textbf{12} & \textit{(8.75)}  & cklwanfifa\textsuperscript{*}           & BSL            & \textit{(11)} & BSL          & \textit{(11)}  & BSL            & \textit{(7)}  & 0.469   & \textit{(6)} \\
            \textbf{12} & \textit{(8.75)} & harikrishnan\_gs\textsuperscript{*}    & 1.295          & \textit{(8)}  & BSL            & \textit{(11)} & BSL            & \textit{(7)}  & BSL   & \textit{(9)} \\
            \textbf{12} & \textit{(8.75)} & tdang\textsuperscript{*}               & BSL            & \textit{(11)} & BSL            & \textit{(11)} & 1.665          & \textit{(4)}  & BSL   & \textit{(9)} \\
            \textbf{15} & \textit{(9.5)}  & \textit{Random baseline (BSL)}& 1.543         & \textit{(11)} & 0.687          & \textit{(11)} & 3.350          & \textit{(7)}  & 0.676 & \textit{(9)} \\
        \hline
    \end{tabular}
\caption{\label{tab:CE_leaderboard} Overall Task A (soft evaluation) results as an average of a system's rank across datasets. \textsuperscript{*} indicates that no system description was available for the team. %(LeWiDi~3).
}
\end{table*}

\begin{table*}[!ht]
    \centering
    \small
    \begin{tabular}{|cr|l|cc|cc|cc|cc|}
        \toprule
        &\multicolumn{10}{c|}{\textbf{\textsc{PERSPECTIVIST EVALUATION }}}\\ 
        \multirow{2}{*}{\textbf{Rank}}&\multirow{2}{0.8cm}{\textbf{(av.pos)}}&\multirow{2}{*}{\textbf{\textsc{Team}}}&\multicolumn{2}{c|}{\textsc{\textbf{CSC}}} &\multicolumn{2}{c|}{\textsc{\textbf{MP}}} &\multicolumn{2}{c|}{\textsc{\textbf{Par}}} &\multicolumn{2}{c|}{\textsc{\textbf{VEN}}} \\ 
        &&&\textbf{MAD}&\textbf{ \textit{(rank)}}&\textbf{ER}&\textbf{ \textit{(rank)}} &\textbf{MAD}& \textbf{ \textit{(rank)}} &\textbf{MER}&\textbf{ \textit{(rank)}} \\ \hline
         \midrule
            \textbf{1}  & \textit{(1.5)}  & Opt-ICL              & \textbf{0.156} & \textit{(1)}  & \textbf{0.289} & \textit{(1)}  & 0.119 & \textit{(2)}  & 0.270 & \textit{(2)} \\
            \textbf{2}  & \textit{(2)}    & DeMeVa              & 0.172          & \textit{(2)}  & 0.300          & \textit{(2)}  & 0.134 & \textit{(2)}  & 0.228 & \textit{(2)} \\
            \textbf{3}  & \textit{(3.25)} & twinhter            & 0.228          & \textit{(5)}  & 0.319          & \textit{(6)}  & \textbf{0.080} & \textit{(1)}  & \textbf{0.124} & \textit{(1)} \\
            \textbf{4}  & \textit{(3.75)} & McMaster            & 0.213          & \textit{(3)}  & 0.311          & \textit{(2)}  & 0.199 & \textit{(4)}  & 0.343 & \textit{(6)} \\
            \textbf{5}  & \textit{(4.75)}    & \textit{Most frequent baseline} & 0.239          & \textit{(5)}  & 0.316          & \textit{(2)}  & 0.362 & \textit{(6)}  & 0.345 & \textit{(6)} \\
            \textbf{6}  & \textit{(5)}  & aadisanghani\textsuperscript{*}        & 0.213          & \textit{(3)}  & 0.311          & \textit{(2)}  & 0.491 & \textit{(6)}  & BSL   & \textit{(9)} \\
            \textbf{6}  & \textit{(5)}  & BoN Appetite Team   & 0.231          & \textit{(5)}  & 0.414          & \textit{(9)}  & 0.228 & \textit{(4)}  & 0.272 & \textit{(2)} \\
            \textbf{8}  & \textit{(6.5)}  & NLP-ResTeam         & 0.291          & \textit{(8)}  & 0.326          & \textit{(6)}  & 0.418 & \textit{(6)}  & 0.345 & \textit{(6)} \\
            \textbf{9}  & \textit{(7)}   & cklwanfifa \textsuperscript{*}         & BSL            & \textit{(10)} & BSL            & \textit{(10)} & BSL   & \textit{(6)}  & 0.271 & \textit{(2)} \\
            \textbf{10} & \textit{(7.5)}   & LPI-RIT             & 0.331          & \textit{(9)}  & 0.324          & \textit{(6)}  & 0.437 & \textit{(6)}  & BSL   & \textit{(9)} \\
            \textbf{11} & \textit{(8.75)}   & \textit{Random baseline (BSL)} & 0.352        & \textit{(10)} & 0.499          & \textit{(10)} & 0.367 & \textit{(6)}  & 0.497 & \textit{(9)} \\
        \hline
    \end{tabular}
\caption{\label{tab:perspectivist_leaderboard} Overall Task B (perspectivist evaluation) results as an average of a system's rank across datasets. \textsuperscript{*} indicates that no system description was available for the team.
}
\end{table*}

\subsection{General discussion}

As in the previous edition of the shared task, we observed a great variety in design choices, but some trends emerge.
\paragraph{System choices}
Some teams (\SYSTEM{OCP-ICL}, \SYSTEM{DeMeVa}, \SYSTEM{BoN Appetit Team}) used large language models relying on in-context learning (ICL) or test-time scaling methods. Others built on transformer models (RoBERTa, BERT, or XLM-R) and trained on the shared task data with annotator-aware extensions (\SYSTEM{McMaster}, \SYSTEM{twinhter}, \SYSTEM{NLP-ResTeam}), or with data augmentation and ensembles but without explicit annotator features (\SYSTEM{PromotionGo}). Finally, hybrid systems included \SYSTEM{LPI-RIT}, which combined sentence-transformer embeddings with the DisCo architecture, and \SYSTEM{Uncertain (Mis)Takes}, which modeled disagreement via semantic entropy over LLMs' generated explanations.
\paragraph{Towards Unified Approaches}
A clear difference from the previous edition (where teams tailored systems to each dataset) is that all participants who submitted for more than one dataset pursued general-purpose pipelines, aiming to capture patterns of disagreement across datasets with a unified approach. The majority instantiates a separate model for each dataset but follows the same pipeline, while others use a single model uniformly for all datasets. 
\paragraph{Overall Rankings and Local Exceptions}
As a consequence of the shift away from dataset-specific solutions toward general-purpose pipelines, a clearer view of which approaches generalize better was enabled.
In fact, differently from the previous edition, some systems ranked consistently among the best across all datasets and tasks. 
LLM-based systems with ICL secured the top positions in the overall leaderboard, with \SYSTEM{OCP-ICL} and \SYSTEM{DeMeVa} ranking first and second. However, fine-tuned transformer models, such as \SYSTEM{twinhter} and \SYSTEM{McMaster} were competitive and \SYSTEM{twinhter} outperformed LLMs on smaller datasets \textit{Par} and \textit{VEN}. Moreover, the specific leaderboards revealed notable exceptions: teams that focused on tailored solutions for a single dataset, \SYSTEM{PromotionGo} on \textit{Par} and \SYSTEM{Uncertain (Mis)Takes} on \textit{VEN}, achieved first place locally.

\paragraph{Annotator information}
The majority of teams (six) used annotator information extensively, %(all except \SYSTEM{Promotion Go,BoN Appetit Team, Uncertain Mis(Takes)})
 devoting effort to find the optimal way for encoding annotator information. Two types of information were available: annotators’ previous behavior and demographics. %Two types of information were available: past annotator's behavior and demographics. 
Some systems used annotator examples in in-context prompts to learn annotator views with LLMs (\SYSTEM{Opt-ICL}, \SYSTEM{DeMeVa}) or implicitly by training on each pair annotation-item or by passing annotator ID (\SYSTEM{twinhter},
\SYSTEM{NLP-ResTeam},  \SYSTEM{LPI-RIT}).
Demographics information usage was tested by \SYSTEM{Opt-ICL}, \SYSTEM{McMaster}, \SYSTEM{twinther} and \SYSTEM{NLP-ResTeam}. Notably, all of the best-performing systems incorporated some form of annotator information. Further details on the impact of annotator information are in Section \ref{par:ablation}.

\paragraph{Data Augmentation Strategies}

\SYSTEM{Opt-ICL} post-trained LLMs using over 40 additional datasets. \SYSTEM{NLP-ResTEAM} synthesized examples via paraphrasing and back-translation. \SYSTEM{PromotionGo} applied extensive lexical (swap and reformulation) and translation-based augmentation. Further details on the impact of data augmentation are given in Section \ref{par:ablation}.

\paragraph{Task A vs Task B}
Leaderboard rankings for the two complementary tasks were largely similar. Not all systems attempted Task B, but of those that did, several  derived the soft labels for Task A from the perspectivist labels for Task B. 
All three top-performing systems adopted this strategy, indicating that understanding annotator behavior contributes to overall prediction quality.
Other systems adopted a multi-task strategy, 
using one output head for the soft label, the other for the perspectivist information.

\subsection{Individual datasets results}\label{sec:Single datasets results}

\paragraph{CSC}
Two major observations stand out regarding \textit{CSC}. The first relates to the role of demographic information. Most participating teams have used annotator information in their systems, regardless of their ranking. However, the winning team (\SYSTEM{Opt-ICL}) reports through an ablation study that using demographic information did not significantly improve their results. This might be because the demographic information provided in \textit{CSC} consists only of gender and age, with missing data, reported by the \SYSTEM{twinther} team. Another observation is related to the importance of fine-tuning. While the most successful teams have used a combination of in-context learning while leveraging annotators information, two of these teams (\SYSTEM{DeMeVa} and \SYSTEM{McMaster}) report that fine-tuning RoBERTa has yielded comparable results to in-context learning with larger models. The winning team (\SYSTEM{Opt-ICL}) also reports that dataset-specific fine-tuning was a crucial contributor to the results.

\paragraph{MP} With respect to the other dataset included in the shared task, \textit{MP} presented and additional challenge due to its multilinguality. This challenge was approached by leveraging pre-trained multilingual backbones (the majority of the teams) and/or by fine-tuning on the multilingual data. While the dataset is very metadata-rich, the top-2 best performing models for both tasks either did not incorporate annotators' sociodemographic data or only noticed a slight improvement when doing so. Fine-tuning was used for most systems. Submissions to Task A showed in general better results (with only two teams performing worse than the most frequent BSL), while only the winning team performed significantly better in Task B; we hypothesize this could be due to the large number of annotators in the dataset.

\paragraph{VEN \& Par} \textit{VEN} and \textit{Par} are two datasets with similar designs: (1) the same four annotators annotated all instances in the corpora, (2) all annotators are required to provide explanations to supplement their annotated labels.
% and (3) these annotations and explanations were later used for validation purposes. 
Due to these design similarities, we observe that the Perspectivist rankings of \textit{Par} and \textit{VEN} are extremely similar, with \texttt{twinhter} ranking first and \texttt{Opt-ICL} and \texttt{DeMeVa} in the tied second place. 
All three systems incorporated explanations into the context and demonstrated that models (both BERT-based ones and LLMs) can leverage this richer textual input to better understand labeling rationales and thus enhance performance.
\texttt{DeMeVa} observed that including explanations in prompts helps better understand individual annotators' preferences, e.g., Ann3 for positive labels in \textit{Par}.
Additionally, \texttt{Uncertain (Mis)Takes }participated only and won first place in the \textit{VEN} Task A using LLM-generated explanations and semantic entropy scores. 
Overall, explanations proved to be a valuable resource, either as explicit input features or as generated reasoning traces, and consistently contributed to stronger performance on datasets in both soft-label and perspectivist evaluations.

\subsection{The new evaluation metrics: an assessment}

The introduction of new evaluation metrics aimed to overcome the limitations of cross-entropy and to provide more reliable measures of model performance across diverse settings, including binary, multilabel, and ordinal-scale datasets based on the Likert scale. In practice, the Manhattan and Wasserstein distances offered intuitive and robust evaluations of soft label predictions, while the Error Rate and Average Normalized Absolute Distance enabled perspectivist assessments that better reflected annotator behavior and label structure.

For the multilabel scenario, evaluation relies on the Mean Absolute Manhattan Distance (MAMD) and the Mean Error Rate (MER).\footnote{Further details are reported in Appendix \ref{App-metrics}.} These metrics have been designed to consider each label dimension independently, while simultaneously capturing the overall structure of label co-occurrence within an instance. By design, partially correct predictions incur a lower penalty than completely incorrect predictions. This allows the evaluation to reflect both the distribution of individual labels across annotators and their joint occurrence within the same instance, providing a nuanced measure of system performance in multilabel settings.

For datasets with ordinal labels (i.e., Likert-type scales), the Average Normalized Absolute Distance (ANAD) and the Average Wasserstein Distance (AWD) explicitly incorporate the ordinal nature of the labels. Unlike simple accuracy-based measures, these metrics penalize predictions proportionally to their deviation from the true label. In this way, systems are penalized less when producing outputs that are closer to the correct ordinal value, even if not exact, thereby providing a more faithful evaluation of performance on ordinal data.

Across all metrics, the lower bound remains consistent, with a score of 0 indicating a perfect match. A limitation, however, is that the upper bound is in some cases dataset-dependent (e.g., for the Wasserstein distance), which prevents direct comparisons across datasets.

%\subsection{Interesting Architectural Aspects}
\subsection{Post-Submission Experiments and Ablation studies}
\label{par:ablation}

Beyond their official submissions, all teams conducted supplementary analyses to gain a deeper understanding of their systems. These ablation studies and evaluations of alternative strategies enriched the competition with valuable insights and underscored the participants’ commitment. The results demonstrated that the effectiveness of different approaches varied across datasets, reflecting both the specific characteristics of the data and the influence of the evaluation metrics employed.

One major focus investigated was the role of annotator information. For LLM-based systems such as \SYSTEM{OCP-ICL} and \SYSTEM{DeMeVa}, provide in-context rater examples at inference time proved decisive: \SYSTEM{OCP-ICL} showed that such examples drove large gains across datasets while demographics had negligible impact, and \SYSTEM{DeMeVa} demonstrated that stratified selection of annotator examples improved consistency over random or similarity-based sampling. In contrast, for fine-tuned transformer-based models, annotator metadata and embeddings were more influential. \SYSTEM{McMaster} found that demographic embeddings, particularly nationality and ethnicity, improved their RoBERTa system; \SYSTEM{twinhter} observed stronger benefits from annotator metadata on small-annotator datasets; \SYSTEM{LPI-RIT} reported that simple annotator ID tokens stabilized predictions; and \SYSTEM{NLP-ResTeam} showed that label-style composite embeddings often outperformed demographics, though the best choice varied depending on the evaluation metric.

Ablation studies across papers revealed mixed effects of augmentation across teams. \SYSTEM{OCP-ICL} found that post-training on over 40 dataset improved results only for \textit{MP}, while for the other datasets was indifferent. \SYSTEM{NLP-ResTeam} concluded that augmentation helped for small datasets (\textit{Par} and \textit{VEN}), while \SYSTEM{PromotionGo} found that combining augmentation strategies worked best.

\section{Conclusions}

We are delighted that the third edition of the {\LEWIDI} shared task continued to attract the attention of the community researching   disagreement and variation in {\NLP}. 
Again, we found that the participating teams engaged actively with the tasks, tackling  interesting issues such as how best to use annotator information and the relation between soft-label modelling and perspectivist modelling.

Our hope is that the shared task and the datasets we released will stimulate further research in this area, by the participant groups and others. 
We believe that further thinking is still needed on issues such as the most appropriate form of evaluation for tasks in which human subjects express ordinal judgments, or the usefulness of modelling individual annotators or groups of annotators. 
To promote this, the \textit{Codabench} page will remain open to submissions after the deadline so that researchers can continue test their models on the datasets.

\section*{Limitations}

While this edition broadened the range of datasets, the scope remained restricted to text, leaving open the question of how disagreement-aware methods would perform in other modalities such as vision, speech, or multimodal tasks. Another open issue is that all annotators present in the test sets were also seen during training and development. As a result, the shared task did not directly evaluate systems’ ability to generalize to unseen annotators, an ability that is likely to be critical in real-world applications.
\section*{Acknowledgments}

The work of E. Leonardelli has been partially supported by the European Union’s CERV fund under grant agreement No.101143249 (HATEDEMICS). 
The work of V. Basile was funded by the ‘Multilingual Perspective-Aware NLU’ project in partnership with Amazon Science. 
M. Pavlovic was supported by a PhD studentship from DeepMind.
B. Plank acknowledges funding by the
ERC Consolidator Grant DIALECT 101043235.
% Not sure why the following was deleted
M. Poesio was partially supported by  NWO through the AINed Fellowship Grant NGF.1607.22.002 ‘Dealing with Meaning Variation in NLP’.

\bibliographystyle{acl_natbib}
\bibliography{LeWiDi2025}

\appendix
\section*{Appendix} 
%\label{sec:appendix}
%\end{table}
\section{Evaluation Metrics for the Multilabel setting} \label{App-metrics}
In this section we outline how the adopted metrics were adapted to handle multilabel classification.% and ordinal-scale settings.

%\subsection{Multilabel Scenario}
\subsection{Multilabel Average Manhattan Distance (MAMD)}

To account for the multilabel setting, the Average Manhattan Distance  (AMD) %(as average of the Manhattan distances of the per-label distributions - Equation \ref{AMD}) 
was adapted into the Multilabel Average Manhattan Distance (MAMD) reported in equation \ref{MAMD}.  For each sample, the average Manhattan distance across all label-specific distributions is computed. The final score is then obtained as the average of such values over all samples.

\begin{equation}
    \label{AMD}
    AMD(i) = \frac{1}{L} \sum_{j=1}^{L} \sum_{k=1}^{n} \left| p_{i,j,k} - t_{i,j,k} \right|
\end{equation}

\begin{equation}
    \label{MAMD}
    MAMD = \frac{1}{N} \sum_{i=1}^{N} AMD(i)
\end{equation}

With:
\begin{itemize}
    \item $N$ is the total number of samples,
    \item $L$ is the number of labels (e.g., Entailment, Neutral, Contradiction for the VEN dataset),
    \item $n$ is the length of each distribution,
    \item $t_{i,j,k}$ is the $k$-th value of the $j$-th target distribution for sample $i$,
    \item $p_{i,j,k}$ is the corresponding predicted value.
\end{itemize}

\subsection{Multilabel Error Rate (MER)}

The metric adopted for the perspectivist evaluation is the Multilabel Error Rate (MER), which quantifies the average dissimilarity between predicted and target label vectors across multiple samples.
The Multilabel Error Rate (MER) is computed as the average of the average Error Rate %(i.e., average of the individual match scores over all $L$ labels distributions) 
values across all samples as shown in Equation \ref{MER}:

\begin{comment}
\begin{equation}
    \label{AER}
    AER(i) = \frac{1}{L} \sum_{j=1}^{L} \left( 1 - \frac{a - \sum_{k=1}^{a} |t_{i,k} - p_{i,k}|}{a} \right)
\end{equation}
\end{comment}

\begin{equation}
\label{MER}
\begin{aligned}
    MER &= \frac{1}{N} \sum_{i=1}^{N} \left( \frac{1}{L} \sum_{j=1}^{L} ER(i)\right)\\
    &= \frac{1}{N} \sum_{i=1}^{N} \left(\frac{1}{L} \sum_{j=1}^{L}  1 - \frac{a - \sum_{k=1}^{a} |t_{i,j,k} - p_{i,j,k}|}{a} \right)
\end{aligned}
\end{equation}

Here,
\begin{itemize}
    \item $N$ is the total number of samples.
    \item $L$ is the number of possible labels (i.e., the number of label-specific vectors to evaluate per sample, such as Entailment, Neutral, Contradiction).
    \item $a$ is the length of a target or predicted vector (i.e., the number of annotators contributing to each label vector).
    \item $t_{i,j,k}$ is the $k$-th element of the $j$-th target vector for sample $i$.
    \item $p_{i,j,k}$ is the $k$-th element of the $j$-th predicted vector for sample $i$. 
\end{itemize}

\section{Datasets specific leaderboards}
\label{appx-leaderboard}

\begin{table}[!ht]
    \centering
    \setlength{\tabcolsep}{1.75pt} 
    \small
    \begin{tabular}{|clc|clc|}    
        \toprule

        \multicolumn{6}{|c|}{\textbf{\textsc{CSC}}}\\ 
        \multicolumn{3}{|c|}{\textbf{\textsc{TASK A}}}&\multicolumn{3}{c|}{\textbf{\textsc{TASK B}}}\\ 
        \textbf{\textsc{}}&{\textsc{\textbf{Team}}}&\textsc{\textbf{WS}} &\textbf{\textsc{}}&{\textsc{\textbf{Team}}}&\textsc{\textbf{MAD}}\\ 
        \hline
         \midrule 
\textbf{1}&Opt-ICL&0.746&\textbf{1}&Opt-ICL&0.156\\
\textbf{1}&DeMeVa&0.792&\textbf{2}&DeMeVa&0.172\\
\textbf{3}&McMaster&0.803&\textbf{3}&McMaster&0.213\\
\textbf{3}&aadisanghani&0.803&\textbf{3}&aadisanghani&0.213\\
\textbf{5}&twinhter&0.835&\textbf{5}&twinhter&0.228\\
\textbf{6}&BoN Appetit Team&0.928&\textbf{5}&BoN Appetit Team&0.231\\
\textbf{7}&\textit{Most frequent BSL}&1.170&\textbf{5}&\textit{Most frequent BSL}&0.239\\
\textbf{8}&harikrishnan\_gs&1.295&\textbf{8}&NLP-ResTeam&0.291\\
\textbf{9}&NLP-ResTeam&1.393&\textbf{9}&LPI-RIT&0.331\\
\textbf{9}&LPI-RIT&1.451&\textbf{10}&\textit{Random label BSL}&0.352\\
\textbf{11}&\textit{Random label BSL}&1.543&&&\\
\hline
            \end{tabular}
    \caption{\label{tab:csc_results} Results for the CSC dataset }
\end{table}

\begin{table}[!ht]
    \centering
    \setlength{\tabcolsep}{1.75pt} 
    \small
    \begin{tabular}{|clc|clc|}    
        \toprule

        \multicolumn{6}{|c|}{\textbf{\textsc{MP}}}\\ 
        \multicolumn{3}{|c|}{\textbf{\textsc{TASK A}}}&\multicolumn{3}{c|}{\textbf{\textsc{TASK B}}}\\ 
        \textbf{\textsc{}}&{\textsc{\textbf{Team}}}&\textsc{\textbf{MD}} &\textbf{\textsc{}}&{\textsc{\textbf{Team}}}&\textsc{\textbf{ER}}\\ 
        \hline
         \midrule 
 \textbf{1}&Opt-ICL&0.422&\textbf{1}&Opt-ICL&0.289\\
 \textbf{1}&PromotionGo&0.428&\textbf{2}&DeMeVa&0.300\\
 \textbf{3}&McMaster&0.439&\textbf{2}&McMaster&0.311\\
 \textbf{3}&aadisanghani&0.439&\textbf{2}&aadisanghani&0.311\\
 \textbf{5}&twinhter&0.447&\textbf{2}&\textit{Most frequent BSL}&0.316\\
 \textbf{6}&BoN Appetit Team&0.466&\textbf{6}&twinhter&0.319\\
 \textbf{6}&DeMeVa&0.469&\textbf{6}&LPI-RIT&0.324\\
 \textbf{8}&\textit{Most frequent BSL}&0.518&\textbf{6}&NLP-ResTeam&0.326\\
 \textbf{9}&LPI-RIT&0.540&\textbf{9}&BoN Appetit Team&0.414\\
 \textbf{9}&NLP-ResTeam&0.551&\textbf{10}&\textit{Random label BSL}&0.499\\
 \textbf{11}&\textit{Random label BSL}&0.687&&&\\
\hline
            \end{tabular}
    \caption{\label{tab:mp_results} Results for the MP dataset }
\end{table}

\begin{table}[!ht]
    \centering
    \setlength{\tabcolsep}{1.75pt} 
    \small
    \begin{tabular}{|clc|clc|}    
        \toprule

        \multicolumn{6}{|c|}{\textbf{\textsc{Par}}}\\ 
        \multicolumn{3}{|c|}{\textbf{\textsc{TASK A}}}&\multicolumn{3}{c|}{\textbf{\textsc{TASK B}}}\\ 
        \textbf{\textsc{}}&{\textsc{\textbf{Team}}}&\textsc{\textbf{WS}} &\textbf{\textsc{}}&{\textsc{\textbf{Team}}}&\textsc{\textbf{MAD}}\\ 
        \hline
         \midrule 
\textbf{1}&twinhter&0.983&\textbf{1}&twinhter&0.080\\
\textbf{1}&DeMeVa&1.120&\textbf{2}&Opt-ICL&0.119\\
\textbf{1}&Opt-ICL&1.200&\textbf{2}&DeMeVa&0.134\\
\textbf{4}&McMaster&1.605&\textbf{4}&McMaster&0.199\\
\textbf{4}&tdang&1.665&\textbf{4}&BoN Appetit Team&0.228\\
\textbf{4}&BoN Appetit Team&1.797&\textbf{6}&\textit{Most frequent BSL}&0.362\\
\textbf{7}&aadisanghani&3.051&\textbf{6}&\textit{Random label BSL}&0.367\\
\textbf{7}&NLP-ResTeam&3.136&\textbf{8}&NLP-ResTeam&0.418\\
\textbf{7}&\textit{Most frequent BSL}&3.231&\textbf{8}&LPI-RIT&0.437\\
\textbf{7}&\textit{Random label BSL}&3.350&\textbf{8}&aadisanghani&0.491\\
\textbf{7}&LPI-RIT&3.715&&&\\
\hline
            \end{tabular}
    \caption{\label{tab:par_results} Results for the Par dataset }
\end{table}

\begin{table}[!ht]
    \centering
    \setlength{\tabcolsep}{1.6pt} 
    \small
    \begin{tabular}{|clc|clc|}    
        \toprule

        \multicolumn{6}{|c|}{\textbf{\textsc{VEN}}}\\ 
        \multicolumn{3}{|c|}{\textbf{\textsc{TASK A}}}&\multicolumn{3}{c|}{\textbf{\textsc{TASK B}}}\\ 
        \textbf{\textsc{}}&{\textsc{\textbf{Team}}}&\textsc{\textbf{MMD}} &\textbf{\textsc{}}&{\textsc{\textbf{Team}}}&\textsc{\textbf{MER}}\\ 
        \hline
         \midrule 
\textbf{1}&twinhter&0.233&\textbf{1}&twinhter&0.124\\
\textbf{1}&Uncertain Mis(Takes)&0.308&\textbf{2}&DeMeVa&0.228\\
\textbf{3}&BoN Appetit Team&0.356&\textbf{2}&Opt-ICL&0.270\\
\textbf{3}&DeMeVa&0.382&\textbf{2}&cklwanfifa&0.271\\
\textbf{3}&Opt-ICL&0.449&\textbf{2}&BoN Appetit Team&0.272\\
\textbf{6}&cklwanfifa&0.469&\textbf{6}&McMaster&0.343\\
\textbf{7}&\textit{Most frequent BSL}&0.595&\textbf{6}&NLP-ResTeam&0.345\\
\textbf{7}&McMaster&0.638&\textbf{6}&\textit{Most frequent BSL}&0.345\\
\textbf{9}&\textit{Random label BSL}&0.676&\textbf{9}&\textit{Random label BSL}&0.497\\
\textbf{10}&NLP-ResTeam&1.000&&&\\

\hline
            \end{tabular}
    \caption{\label{tab:ven_results} Results for the VEN dataset }
\end{table}

\end{document}